\documentclass[10pt,twocolumn,letterpaper]{article}
\usepackage{cvpr}              

\usepackage{graphicx}
\usepackage{amsmath}
\usepackage{amssymb}
\usepackage{booktabs}
\usepackage{CJKutf8}
\usepackage{soul}
\usepackage[pagebackref,breaklinks,colorlinks]{hyperref}

\usepackage[capitalize]{cleveref}
\crefname{section}{Sec.}{Secs.}
\Crefname{section}{Section}{Sections}
\Crefname{table}{Table}{Tables}
\crefname{table}{Tab.}{Tabs.}

\def\cvprPaperID{3621} 
\def\confName{CVPR}
\def\confYear{2022}

\begin{document}

\title{Aesthetic Text Logo Synthesis via Content-aware Layout Inferring}

\author{Yizhi Wang$^1$\thanks{This work was partly done when Yizhi Wang was an intern in Tencent }, Guo Pu$^1$, Wenhan Luo$^2$, Yexin Wang$^2$, Pengfei Xiong$^2$, Hongwen Kang$^2$, Zhouhui Lian$^1$\thanks{Corresponding author. E-mail: lianzhouhui@pku.edu.cn}\\
$^1$Wangxuan Institute of Computer Technology, Peking University, China\\
$^2$PCG, Tencent\\}

\maketitle
\begin{abstract}
Text logo design heavily relies on the creativity and expertise of professional designers, in which arranging element layouts is one of the most important procedures. However, few attention has been paid to this task which needs to take many factors (e.g., fonts, linguistics, topics, etc.) into consideration.
In this paper, we propose a content-aware layout generation network which takes glyph images and their corresponding text as input and synthesizes aesthetic layouts for them automatically. Specifically, we develop a dual-discriminator module, including a sequence discriminator and an image discriminator, to evaluate both the character placing trajectories and rendered shapes of synthesized text logos, respectively. Furthermore, we fuse the information of linguistics from texts and visual semantics from glyphs to guide layout prediction, which both play important roles in professional layout design. To train and evaluate our approach, we construct a dataset named as TextLogo3K, consisting of about 3,500 text logo images and their pixel-level annotations. Experimental studies on this dataset demonstrate the effectiveness of our approach for synthesizing visually-pleasing text logos and verify its superiority against the state of the art.
\end{abstract}

\section{Introduction}
\label{sec:intro}
\begin{figure}[t!]
  \centering
  \includegraphics[width=\columnwidth]{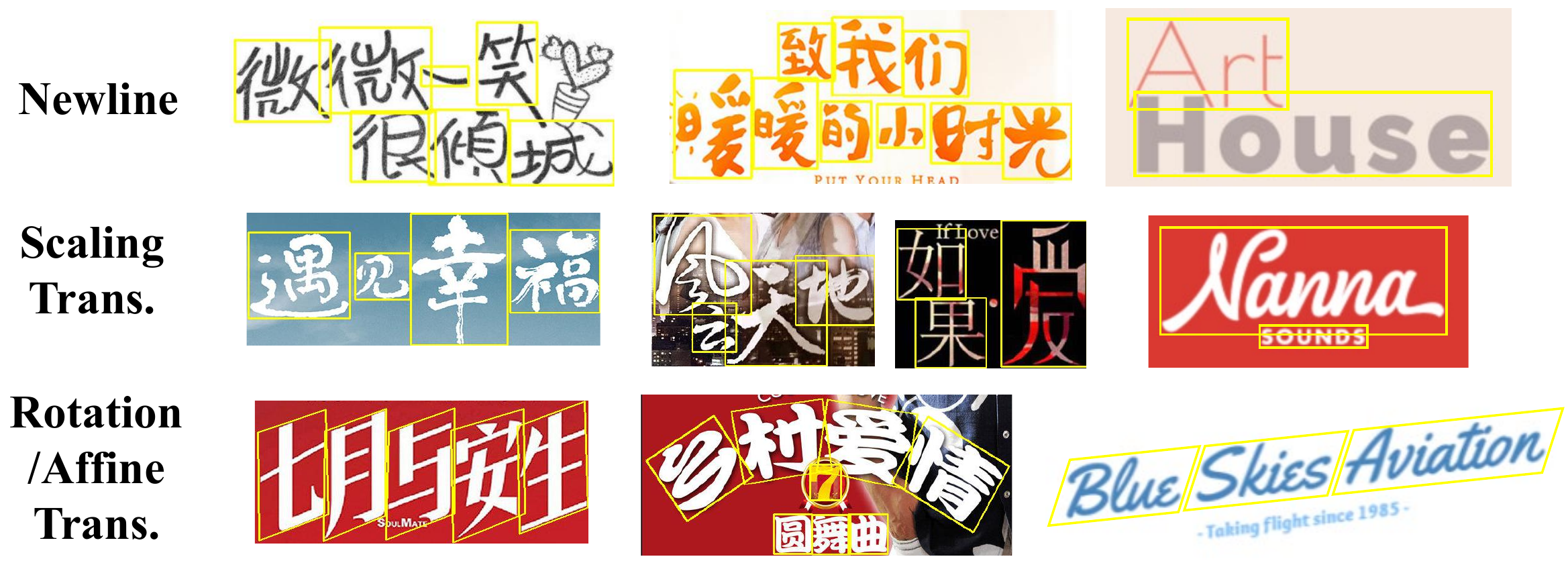}
  \vspace{-0.6cm}
  \caption{Examples of some commonly seen layout types for text logos. ``Trans.'' denotes transformation. Yellow boxes denote the bounding boxes of characters/words.}
  \label{fig:LogoLayoutClasses}
\end{figure}

Creative AI has attracted considerable interests in recent years.
Automatic layout design is a promising and valuable technique, which is quite helpful to facilitate the tasks of designers and improve the efficiency of creating media content. One intuitive way to achieve this goal is to generate layouts by defining specific easy-to-execute rules, but the results are typically plain and without design aesthetics.
Therefore, various methods~\cite{zheng2019content,li2019layoutgan,jyothi2019layoutvae,lee2020neural,li2020attribute,guo2021vinci,arroyo2021variational,Yang_2021_CVPR,Gupta_2021_ICCV} based on deep generative models (such as GAN~\cite{goodfellow2014generative} and VAE~\cite{kingma2013auto}) have been proposed recently to synthesize graphic layouts by learning from human-designed data. 
Text logo design is a challenging task because many factors need to be taken into consideration, such as fonts, layouts, and textures.
As we know, the layout of a text logo is closely related to the logo's text and the selected font styles. 
However, existing approaches~\cite{li2019layoutgan,zheng2019content,guo2021vinci,li2020attribute} exploit only the visual, category, and topic (keyword) information of elements, or combine some of them to generate the layouts.
In fact, the semantics of texts are also very important to determine the layout design. As shown in Figure~\ref{fig:LogoLayoutClasses}, in a desired layout, the Chinese character \begin{CJK}{UTF8}{gbsn}``爱"\end{CJK} (``love") is a verb and the key character in the text logo, so it is emphasized by a larger font size than other characters such as \begin{CJK}{UTF8}{gbsn}``如果"\end{CJK} (``if"), which is a less-important conjunction word.
Moreover, the synthesized layouts need to take fine-grained details into consideration, such as avoiding the collision of strokes of different glyphs.
In addition, the placing trajectories of characters should follow a correct order for reading (e.g., left to right and up to bottom for English) and possess diverse styles at the same time, which cannot be easily handled by non-sequence generation models.
\par
In this paper, we cast the layout synthesis of text logos as a sequence generation problem and employ a GAN-based model\footnote{Code and dataset: https://github.com/yizhiwang96/TextLogoLayout} to solve it.
In our approach, a dual-modality fusion scheme is proposed to encode both the visual and linguistic cues of the input glyphs/text as the condition of GAN. It is widely acknowledged that using GAN to generate discrete sequences is difficult. However, the geometric parameters of design elements are continuous in our task, making GAN training on the sequence of geometric parameters feasible.
As mentioned above, fine-grained details such as glyph collisions and character placing trajectories are critical for the aesthetic quality of a text logo. To this end, a dual-discriminator module is proposed to capture both the character placing trajectory and rendered shape of the synthesized text logo. The placing trajectories of characters are supervised by a sequence discriminator, to guarantee that (1) they follow the correct reading order; and (2) they possess various styles like the human-designed ones.
Moreover, by proposing a differentiable composition method, we build a bridge between the synthesized layouts and rendered text logo images.
At the same time, an image discriminator is also employed to capture the detailed structures of the synthesized text logos, whose visual quality can be further improved in this manner.

\par
To train and evaluate our model, a large-scale dataset of text logos, TextLogo3K, is constructed with pixel-wise annotations. Extensive experiments including ablation studies and method comparison have been conducted on the new dataset. Experimental results clearly demonstrate the effectiveness and advantages of our proposed method. We also demonstrate how our proposed model can be integrated with font generation~\cite{wang2020attribute2font} and texture transfer~\cite{men2019dyntypo} models to automatically synthesize visually-pleasing text logos. To the best of our knowledge, this work presents the first text logo synthesis method that is capable of taking both linguistic (i.e., text contents) and visual (i.e., glyph shapes, font styles, and textures) information into account like professional human designers. Furthermore, our TextLogo3K is also the first dataset that specifically targets at the task of text logo synthesis, which will be released to promote the investigation of this topic and other relevant tasks in the community.

\section{Related Work}
\noindent\textbf{GANs}. In recent years, Generative Adversarial Networks (GANs)~\cite{goodfellow2014generative} have achieved great success in the area of image synthesis.
Mirza et al.~\cite{mirza2014conditional} proposed Conditional GAN by adding the data label as a parameter to the input of the generator and discriminator to make the network generate the desired data output. 
GANs have been employed in various tasks like image SR \cite{ledig2017photo}, deblurring \cite{kupyn2018deblurgan,nah2017deep}, dehazing \cite{ren2018gated,li2018single}, text to image translation \cite{reed2016generative,isola2017image} and video prediction \cite{vondrick2016generating,liang2017dual}.


\noindent\textbf{Graphic Layout Generation}. Li et al.~\cite{li2019layoutgan} proposed a GAN-based model equipped with wireframe discriminators, where
Differentiable Wireframe Rendering is utilized to rasterize the input
graphic elements into 2D wireframe images.
Zheng et al.~\cite{zheng2019content} designed a conditional GAN model to arrange the layouts of paragraphs and images for posters by encoding the input images, keywords, and attributes.
Guo et al.~\cite{guo2021vinci} developed a VAE-based model for generating the locations of design elements (e.g., background images, embellishments, and text styles). More recently, Lee et al.~\cite{lee2020neural} proposed a GCN-based model to generate layouts according to both the design components and the specified constraints. Li et al.~\cite{li2020attribute} proposed attribute-conditioned layout GAN to synthesize layouts according to user-specified layout attributes (such as expected area, aspect ratio and reading order).\par
Unfortunately, none of them can properly handle our task. 
Instead of using the wireframe images~\cite{li2019layoutgan}, we propose to rasterize glyphs directly to the canvas so that the discriminator can capture more details from the inputs (such as strokes).
The reading order in \cite{li2020attribute} is rule-based and predefined while our method generates content-adaptive reading orders.
Through experiments, we also find that sequence generative models are more appropriate than CNNs (such as \cite{zheng2019content}) to generate the location of each character.

\noindent\textbf{Sequence Modeling}. Recurrent Neural Networks (RNNs) such as LSTM~\cite{hochreiter1997long} and GRU~\cite{cho2014learning} have been very successful in NLP. Recently, Transformers~\cite{vaswani2017attention} become popular and outperform RNNs when trained with huge amounts of data.
As suggested in Vinci~\cite{guo2021vinci}, sequential ordering is important for designing.
For our task, the position of each glyph image in a text logo should follow the corresponding reading order. 
Therefore, our task can be viewed as a sequence generation problem where a set of coordinates is generated for one design element at each step.

\noindent\textbf{Differentiable Composition}. Reddy et al.~\cite{reddy2020discovering} proposed a differentiable composition method which can find
the type, position, orientation, and layering of elements that form a given pattern image.
However, their method did not deal with the scaling transformation, which is important in layout design.
In this paper, we transform each glyph image into the predicted position on the canvas via a variant of Spatial Transform Networks (STN)~\cite{jaderberg2015spatial}.

\section{Method}
\subsection{Preliminaries}
\begin{figure}[t!]
  \centering
  \includegraphics[width=\columnwidth]{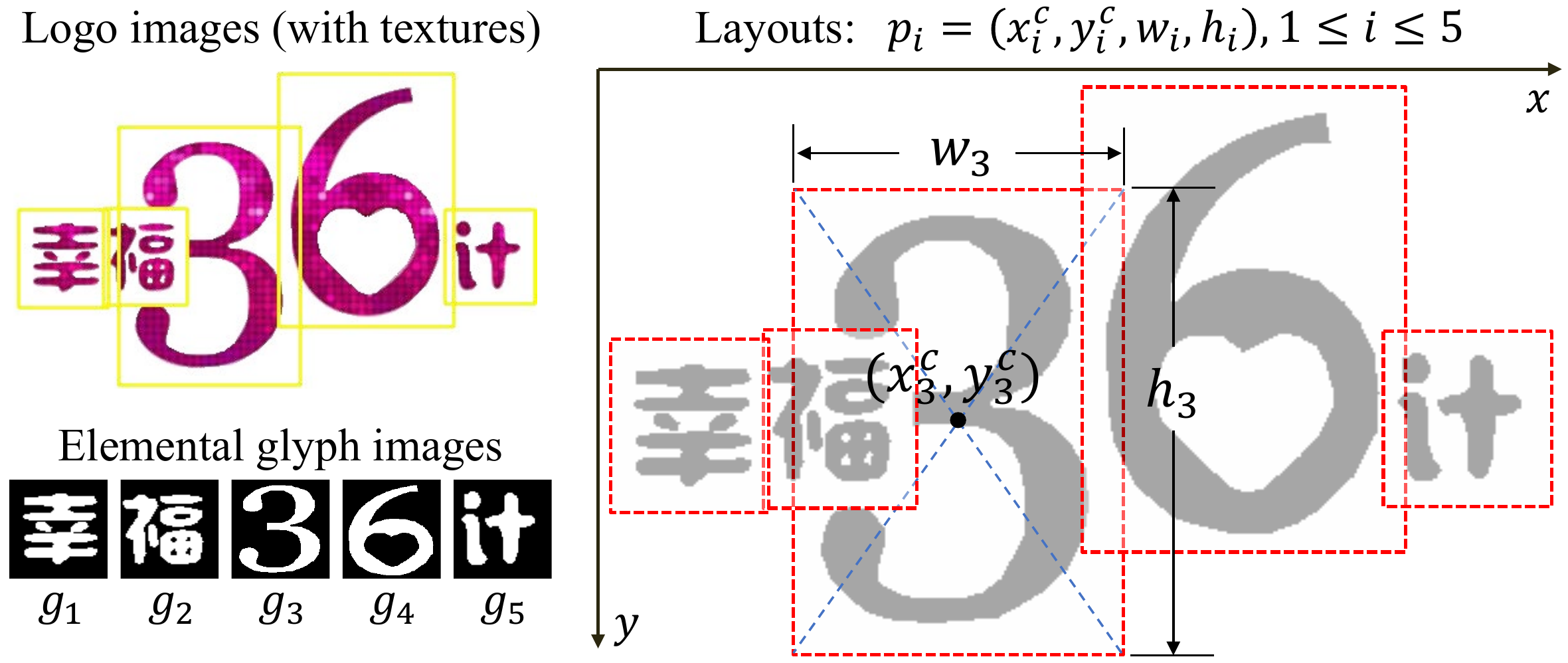}
  \vspace{-0.6cm}
  \caption{An illustration of the layout and elemental glyph images of a text logo image.}
  \label{fig:DataStructure}
\end{figure}

As shown in Figure~\ref{fig:DataStructure}, a text logo is comprised of a set of $N$ glyph elements: ${(g_{1},p_{1}), \cdot \cdot \cdot, (g_{N},p_{N})}$, where $g_{i} \in \mathbb{R}^{H_{g} \times W_{g}}$ is the raw glyph image, $W_{g}$ and $H_{g}$ are the width and height of the glyph image, respectively, and $p_{i}$ denotes the geometric parameters of $g_{i}$ being presented on the canvas.
We assume that the bounding boxes of glyphs are rectangles, \emph{i.e.}, $p_{i} = (x_{i}^{c}, y_{i}^{c}, w_{i}, h_{i})$, where $(x_{i}^{c}, y_{i}^{c})$ are the coordinates of the centre point, $w_{i}$ and $h_{i}$ are the width and height of the rectangle, respectively; $x_{i}^{c}$, $w_{i}$ $\in$ $\left(0,W_{c}\right)$ and $y_{i}^{c}$, $h_{i}$ $\in$ $\left(0,H_{c}\right)$, where $H_{c}$ and $W_{c}$ are the height and width of the logo canvas, respectively.
\par
For our layout generation network, the input consists of the logo's text and its corresponding glyph images $\boldsymbol{g} = (g_{1}, \cdot \cdot \cdot, g_{N})$. We represent the input text by the character embeddings provided by~\cite{li2018analogical}, which are denoted as $\boldsymbol{f}^{e} = (f^{e}_{1}, f^{e}_{2}, \cdot \cdot \cdot, f^{e}_{N})$. The ground-truth (human-designed) layout is denoted as $\boldsymbol{p} = (p_{1}, \cdot \cdot \cdot, p_{N})$.
\subsection{Overview}
\begin{figure*}[t!]
  \centering
  \includegraphics[width=\textwidth]{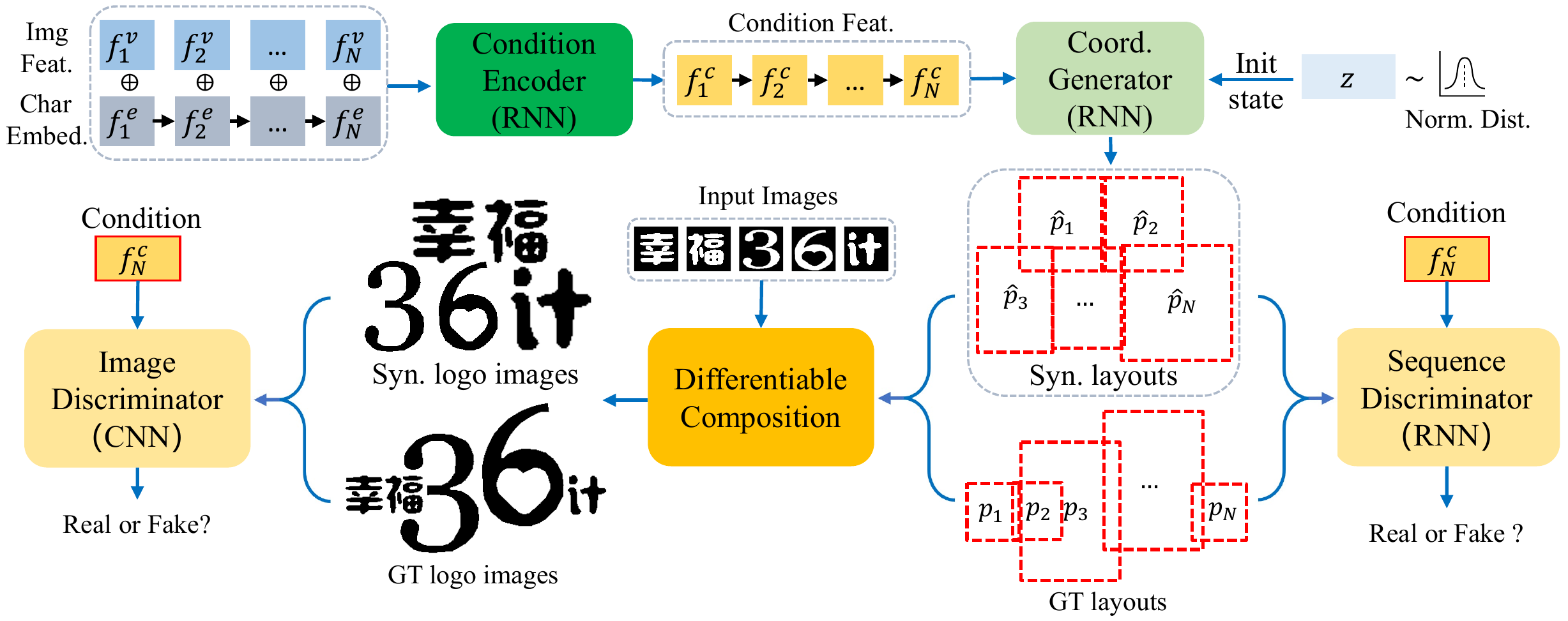}
  \vspace{-0.6cm}
  \caption{The pipeline of our model. ``Char Embed.'' denotes char embeddings, ``Img Feat.'' denotes image features, ``Syn.'' denotes synthesized, ``GT'' denotes ground-truth, $\oplus$ denotes the concatenation operation, and the blue arrow represents the direction of data flow.}
  \label{fig:Pipeline}
\end{figure*}

We illustrate the pipeline of our layout generation model in Figure~\ref{fig:Pipeline} and give more details in the following subsections.
Specifically, we exploit the dual-modality features of input elements (\emph{i.e.}, visual features $\boldsymbol{f}^{v}$ and character embeddings $\boldsymbol{f}^{e}$) and encode them as the condition features $\boldsymbol{f}^{c}$.
The \textbf{Coordinate Generator} takes the condition vector $\boldsymbol{f}^{c}$ and a random noise $z$ as inputs, to predict the geometric parameters of each character that are denoted as $\boldsymbol{ \hat{p}} = (\hat{p}_{1}, \hat{p}_{2}, \cdot \cdot \cdot, \hat{p}_{N})$, where $\hat{p}_{i} = (\hat{x}_{i}^{c}, \hat{y}_{i}^{c}, \hat{w}_{i}, \hat{h}_{i})$ and the four parameters denote the centre coordinates, width and height of the predicted bounding box of $g_{i}$, respectively.
$(\hat{p}_{1}, \hat{p}_{2}, \cdot \cdot \cdot, \hat{p}_{N})$ can be viewed as a placing trajectory of the character sequence, so we employ an RNN-based \textbf{Sequence Discriminator} to distinguish between $\hat{\boldsymbol{p}}$ and $\boldsymbol{p}$ under the condition of $f^{c}_{N}$ (the final state of $\boldsymbol{f}^{c}$).
By performing \textbf{Differentiable Composition} (denoted as $F_{C}$), we have the synthesized text logo image $\hat{l} = F_{C}(\boldsymbol{g},\hat{\boldsymbol{p}})$.
To capture finer details in the logo images and further ensure that the generated images follow the fundamental principles of human design (\emph{e.g.}, there should be no large-portion overlap between characters in text logos), an \textbf{Image Discriminator} is introduced to distinguish between $\hat{l}$ and $l$ under the condition of $f^{c}_{N}$.
\subsection{Encoding Visual and Linguistic Information}
We take both the visual and linguistic information of inputs into consideration to predict the geometric parameters of each glyph.
First, we utilize a CNN network as the visual encoder to extract the visual feature of each glyph image, denoted as $\boldsymbol{f}^{v} = (f^{v}_{1}, f^{v}_{2}, \cdot \cdot \cdot,  f^{v}_{N})$.
Then, we concatenate the visual feature and the character embedding of each input element to compose a new sequence $\boldsymbol{f}^{\prime}$ as $(\left[f^{v}_{1},f^{e}_{1}\right], \left[f^{v}_{2},f^{e}_{2}\right], \cdot \cdot \cdot, \left[f^{v}_{N},f^{e}_{N}\right])$,
where the square bracket denotes concatenation.
Finally, $\boldsymbol{f}^{\prime}$ is sent into a condition encoder to obtain the condition features, denoted as $\boldsymbol{f}^{c} = (f^{c}_{1}, f^{c}_{2},\cdot \cdot \cdot, f^{c}_{N})$.
The state at the last step $f^{c}_{N}$ can be viewed as a holistic conditional representation of the inputs (images and texts), which is utilized as the condition of dual discriminators, including both the sequence discriminator and the image discriminator. The whole encoding process can be defined as $\boldsymbol{f}^{c} = E(\boldsymbol{f}^{v}, \boldsymbol{f}^{e})$.
Afterwards, we fed the condition features $\boldsymbol{f}^{c}$ and a noise $z$ to the coordinate generator $G$, to derive the geometric parameters $\hat{\boldsymbol{p}}$ as $G(\boldsymbol{f}^{c}, z)$,
where $G(\cdot, \cdot)$ is the coordinate generator approximated by an RNN encoder-decoder; $z$ is randomly sampled from the Standard Normal Distribution and set as the initial state of $G$'s encoder;
$\boldsymbol{f}^{c}$ is the input sequence for $G$.
The function of $G$'s encoder is to propagate the layout style information $z$ to each position of the sequence.
The last layer of the coordinate generator is appended with a Sigmoid function whose output is multiplied by $W_{c}$ or $H_{c}$, to guarantee that the predicted parameters are safely in the range of $(0, W_{c})$ for $\hat{x}_{i}^{c}$ and $\hat{w}_{i}$, and $(0, H_{c})$ for $\hat{y}_{i}^{c}$ and $\hat{h}_{i}$.

\subsection{Differentiable Composition}
After obtaining the predicted geometric parameters corresponding to the glyph images, it is required to transform the glyph images into a text logo with the layout according to the glyphs' geometric parameters. More importantly, the transformation process should not destruct the end-to-end optimization process, \emph{i.e.}, the transformation should propagate gradients. To this end,
we render the glyph images on the canvas according to the generated coordinates in a differentiable way on the basis of a variant of STN~\cite{jaderberg2015spatial}.
To ease the illustration, we first briefly introduce the vanilla STN.
Given the source coordinates $(x_{j}^{s},y_{j}^{s})$ of grids in the input feature map, STN aims to learn a set of affine transformation parameters $\Theta$ and transform $(x_{j}^{s},y_{j}^{s})$ into the target coordinates $(x_{j}^{t},y_{j}^{t})$ in the output feature maps that are expected to facilitate the down-stream tasks. The transformation is formulated as
\begin{equation}
\begin{pmatrix}x_{j}^{s}\\y_{j}^{s}\\1\end{pmatrix}=
\Theta \begin{pmatrix}x_{j}^{t}\\y_{j}^{t}\\1\end{pmatrix}=
\begin{bmatrix}\Theta_{11}&\Theta_{12}&\Theta_{13}\\\Theta_{21}&\Theta_{22}&\Theta_{23}\end{bmatrix}\begin{pmatrix}x_{j}^{t}\\y_{j}^{t}\\1\end{pmatrix},
\end{equation}
where $j$ is the index of a grid in the feature maps.
With the established mapping, STN performs Differentiable Image Sampling (DIS) implemented by the bilinear interpolation to obtain the transformed feature maps.
\par
Our task is different from the vanilla STN in that we have already had the predicted coordinates $\hat{p}_{i} = (\hat{x}_{i}^{c}, \hat{y}_{i}^{c}, \hat{w}_{i}, \hat{h}_{i})$ of 
target glyphs. We need to obtain the affine transformation parameters and then perform DIS to derive the final rendered glyphs on the canvas.
In our scenario, only translation and scaling are considered, so we set $\Theta_{12} = 0$ and $\Theta_{21} = 0$.
We use the coordinates of the four corners of the input glyph images ($(0,0)$, $(W_{g},0)$, $(W_{g},H_{g})$, $(0,H_{g})$) and the coordinates of the corresponding transformed images ($(\hat{x}_{i}^{c}-\frac{\hat{w}_{i}}{2},\hat{y}_{i}^{c}-\frac{\hat{h}_{i}}{2})$, $(\hat{x}_{i}^{c}+\frac{\hat{w}_{i}}{2},\hat{y}_{i}^{c}-\frac{\hat{h}_{i}}{2})$, $(\hat{x}_{i}^{c}+\frac{\hat{w}_{i}}{2},\hat{y}_{i}^{c}+\frac{\hat{h}_{i}}{2})$, $(\hat{x}_{i}^{c}-\frac{\hat{w}_{i}}{2},\hat{y}_{i}^{c}+\frac{\hat{h}_{i}}{2})$) to construct the equation set.
Intuitively, the equation set can be theoretically resolved, and the transformation parameters are derived as
\begin{equation}
\Theta_{i}=
\begin{bmatrix} \frac{2 \cdot W_{g}}{\hat{w}_{i}} & 0 &  \frac{(\hat{w}_{i} - 2 \cdot \hat{x}^{c}_{i}) W_{g}}{\hat{w}_{i}} \\
0 &\frac{2 \cdot H_{g}}{\hat{h}_{i}} &  \frac{(\hat{h}_{i} - 2 \cdot \hat{y}^{c}_{i}) H_{g}}{\hat{h}_{i}}\end{bmatrix}\cdot 
\end{equation}
Having the transformation parameters $\Theta$, we apply the corresponding transformation on each glyph image as
\begin{equation}
 g^{\prime}_{i} =F_{A}(g_{i}, \Theta_{i}),
\end{equation}
where $g^{\prime}_{i}$ is the transformed glyph image of $g_{i}$ and $F_{A}$ denotes the transformation function for images implemented as DIS.
Typically, the shapes of different glyphs have no overlap in text logos so we do not need to take the layering of elements into consideration. Therefore, we can perform simple additions to obtain the final logo image $\hat{l} \in \mathbb{R}^{H_{c} \times W_{c}}$ as
\begin{equation}
\hat{l} =  min(\sum_{i=1}^{N} g^{\prime}_{i}, v_{max}),
\end{equation}
where $H_{c}$ and $W_{c}$ are the height and width of the logo canvas, respectively. The output pixel values greater than $v_{max}$ are truncated, to guarantee that the pixel values of $\hat{l}$ are in the range of $[0,v_{max}]$ (typically $v_{max}=255$).
An illustration of transformed glyph images and the obtained logo image is shown in Figure~\ref{fig:OverlapLoss}.

\subsection{The Dual Discriminators}
Character placing trajectories are supposed to conform with the reading order and possess diverse styles.
However, these characteristics cannot be easily captured by CNNs that are commonly used as the discriminator in image synthesis models. To tackle this, a dual-discriminator module including a Sequence Discriminator and an Image Discriminator is proposed.

The Sequence Discriminator $D_{s}$ takes the sequence of geometric parameters ($\boldsymbol{p}$ or $\hat{\boldsymbol{p}}$) as inputs and the encoded condition $f^{c}_{N}$ as the initial state, to analyze the reasonableness of the character placing trajectory. The outputs are denoted as $D_{s}(\hat{\boldsymbol{p}}, f^{c}_{N})$ and $D_{s}(\boldsymbol{p}, f^{c}_{N})$, representing the probabilities of $\hat{\boldsymbol{p}}$ and $\boldsymbol{p}$ being real, respectively.\par
The Sequence Discriminator is unable to capture fine-grained information (such as strokes) because they receive only the geometric parameters as inputs. 
Thus, the Image Discriminator $D_{i}$ is introduced to investigate the details of logo images (synthesized or human-designed) and then predict whether they are real or fake.
Following the suggestion of~\cite{perarnau2016invertible}, we tile the condition feature $f^{c}_{N}$ after the first convolution layer of $D_{i}$.
The outputs of $D_{i}$ are denoted as $D_{i}(\hat{l}, f^{c}_{N})$ and $D_{i}(l, f^{c}_{N})$, representing the probabilities of $\hat{l}$ and $l$ being real, respectively.
\par

\subsection{The Overlap Loss}
\begin{figure}[t!]
  \centering
  \includegraphics[width=\columnwidth]{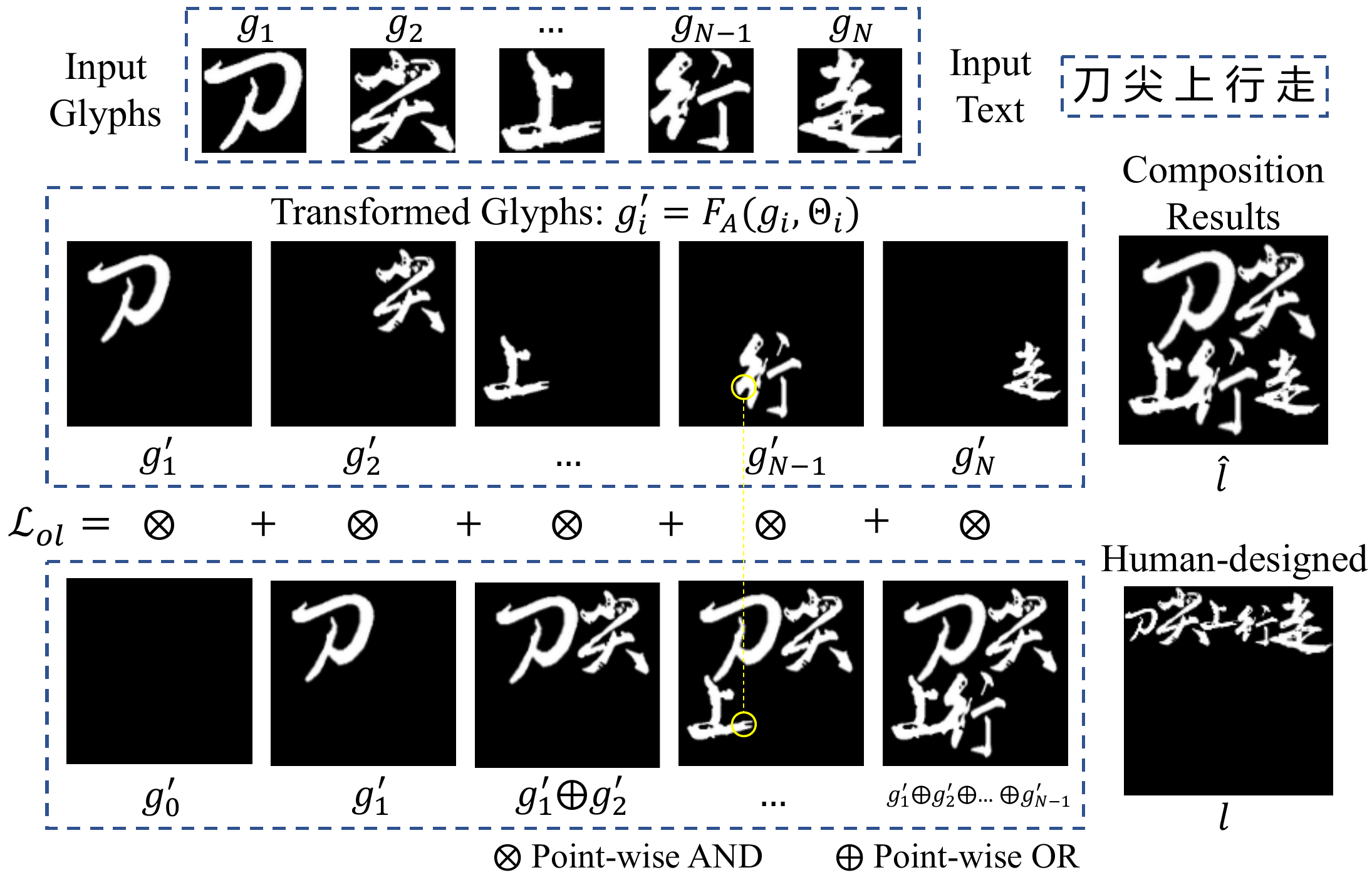}
  \vspace{-0.6cm}
  \caption{An illustration of transforming elemental glyph images into the canvas and how the overlap loss is calculated. The yellow circles highlight the overlap of different glyphs.}
  \label{fig:OverlapLoss}
\end{figure}

Generally, the overlap of different glyphs will affect the readability of the synthesized logos and make them look uncoordinated.
Through our experiments, we discover that small-portion overlaps cannot be distinguished by the Image Discriminator.
Therefore, an overlap loss $\mathcal{L}_{ol}$ is introduced to explicitly penalize the overlap of the transformed glyphs, which is formulated as
\begin{equation}
\mathcal{L}_{ol} = \sum_{i=1}^{N} (g^{\prime}_{i} \otimes (g^{\prime}_{0} \oplus \cdot \cdot \cdot \oplus g^{\prime}_{i-1}) ),
\end{equation}
where $\otimes$ denotes the point-wise AND operation and $\oplus$ denotes the point-wise OR operation; $g^{\prime}_{0}$ is a blank mask which can be omitted.
An example to illustrate the overlap loss is shown in Figure~\ref{fig:OverlapLoss},
where the input text is \begin{CJK}{UTF8}{gbsn}``刀尖上行走"\end{CJK}.
The transformed glyph $g^{\prime}_{4}$ (\begin{CJK}{UTF8}{gbsn}``行"\end{CJK}) overlaps with the previously rendered glyphs ($g^{\prime}_{1} \oplus g^{\prime}_{2} \oplus g^{\prime}_{3})$, which will activate the penalty function $\mathcal{L}_{ol}$.

\subsection{Loss Function}
The loss functions for the Sequence and Image Discriminators are:
\begin{equation}
\begin{aligned}
\mathcal{L}_{D}^{s} =  \log(D_{s}(\boldsymbol{p}, f^{c}_{N})) + \log(1 - D_{s}(\hat{\boldsymbol{p}}, f^{c}_{N})) \\
=  \log(D_{s}(\boldsymbol{p}|E)) + \log(1 - D_{s}(G(E,z)|E)),
\end{aligned}
\end{equation}
\begin{equation}
\begin{aligned}
\mathcal{L}_{D}^{i} =  \log(D_{i}(l, f^{c}_{N})) + \log(1 - D_{i}(\hat{l}, f^{c}_{N})) \\
=  \log(D_{i}(l|E)) + \log(1 - D_{i}(F_{C}(\boldsymbol{g}, G)|E)),
\end{aligned}
\end{equation}
respectively, where we omit the inputs of some functions (such as $E$ and $G$) for brevity.
The overall objective function is
\begin{equation}
\min \limits_{E, G}\max \limits_{D_{s},D_{i}} (\mathcal{L}_{D}^{s} + \mathcal{L}_{D}^{i} + \lambda \mathcal{L}_{ol}),
\end{equation}
where the aim of optimizing the parameters of $E$ and $G$ is to synthesize a realistic sequence $\hat{\boldsymbol{p}}$ and its corresponding logo image $\hat{l}$ to fool the Sequence and Image Discriminators, respectively, while the parameters of $D_{s}$ and $D_{i}$ are optimized to distinguish $\hat{\boldsymbol{p}}$ and $\hat{l}$ from the real ones. $\lambda$ is a hyper-parameter and experimentally chosen.

\section{Dataset}
\begin{figure}[t!]
  \centering
  \includegraphics[width=\columnwidth]{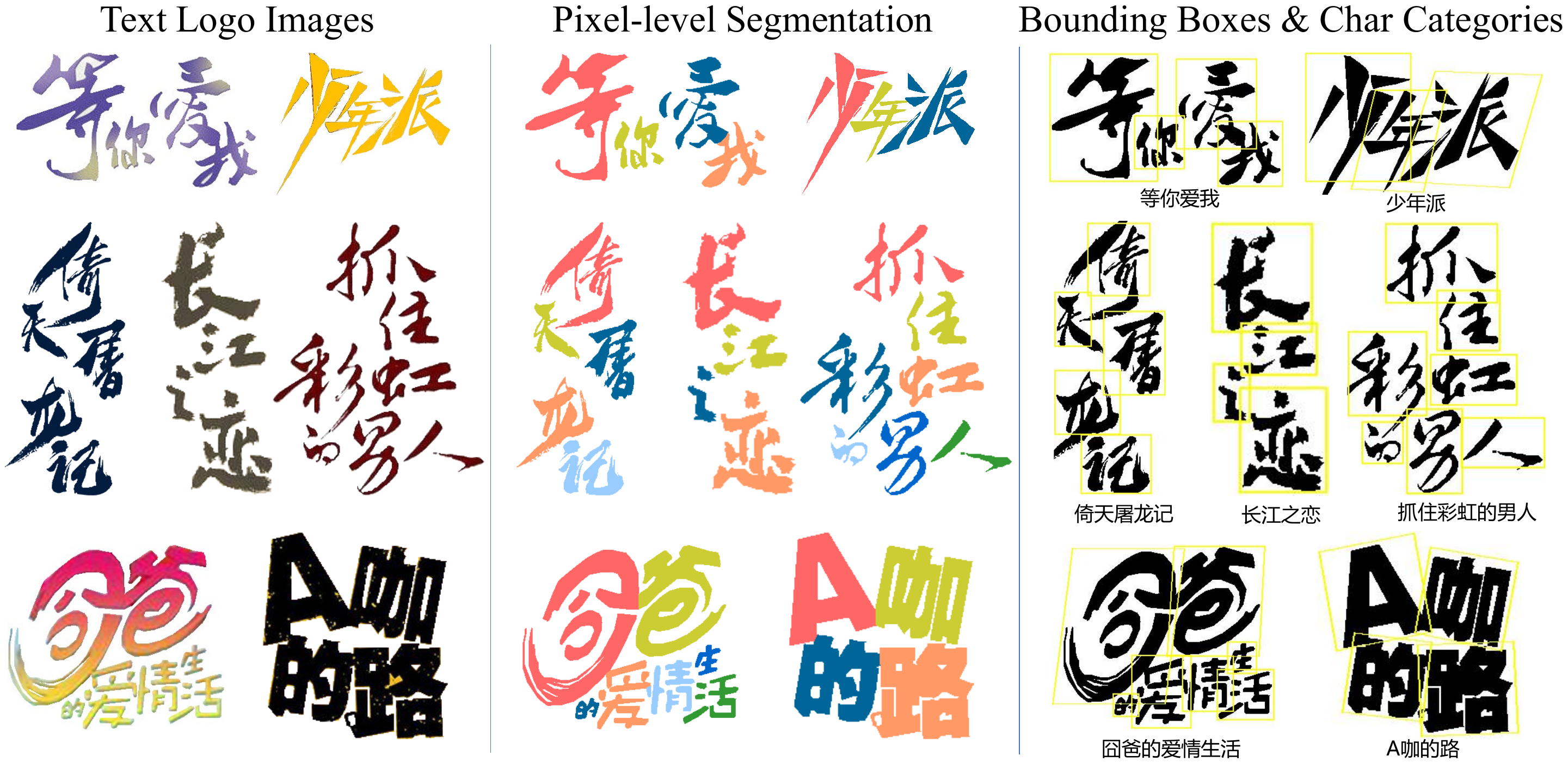}
  \vspace{-0.6cm}
  \caption{The proposed text logo dataset, consisting of 3,470 text logo images with various styles and annotated with pixel-level segmentation, bounding boxes, and character categories.}
  \label{fig:DatasetIntro}
\end{figure}

We construct a text logo dataset named as \textbf{TextLogo3K} by collecting data from Tencent Video, one of the leading online video platforms in China.
The dataset consists of $3,470$ carefully-selected text logo images that are extracted from the posters/covers of the movies, TV series, and comics. 
We manually annotate the bounding box, pixel-level mask, and category for each character in those text logos.
Some examples are shown in Figure~\ref{fig:DatasetIntro}. For each example, a color sequence is used to distinguish between different glyphs in the pixel-level annotations of text logos.
In the bounding box annotations, we only annotate the most centre parts if the glyph has very long strokes. We also annotate the angle if the glyph is rotated or affine-transformed. It is notable that, the rotation or affine transformation are not considered in our model, but it can be applied in these situations with minor modifications.
Besides the layout generation and text logo synthesis tasks, the proposed dataset can also benefit other relevant research (text segmentation, artistic text recognition, artistic font generation, etc.) in the community.\par 
To verify the generality of our model in other writing systems, we make use of the TextSeg dataset proposed in~\cite{xu2021rethinking}, which consists of English text images with pixel-level segmentation. For this dataset, we set the design element as an English word.

\section{Experiments}
\subsection{Implementing Details}
The height $H_{g}$ and width $W_{g}$ of input glyph images are set to 64.
The height $H_{c}$ and width $W_{c}$ of the logo canvas are set to 128.
The maximum number of $N$ is 20.
We split our TextLogo3K dataset into two parts (90\% for training and 10\% for testing).
The visual encoder is a pretrained VGG-19~\cite{simonyan2014very} network which is fine-tuned during training.

\subsection{Evaluation Metrics}
We adopt two commonly-used metrics for image generation to evaluate the performance of our model and others: Fréchet Inception Distance (FID)~\cite{heusel2017gans} and Inception Score (IS)~\cite{salimans2016improved}.
The logo images rendered with our synthesized layouts (viewed as domain A) and the ground-truth logo images from the whole dataset (viewed as domain B) are compared to calculate FID and IS.

\subsection{Synthesized Examples}
\begin{figure}[t!]
  \centering
  \includegraphics[width=\columnwidth]{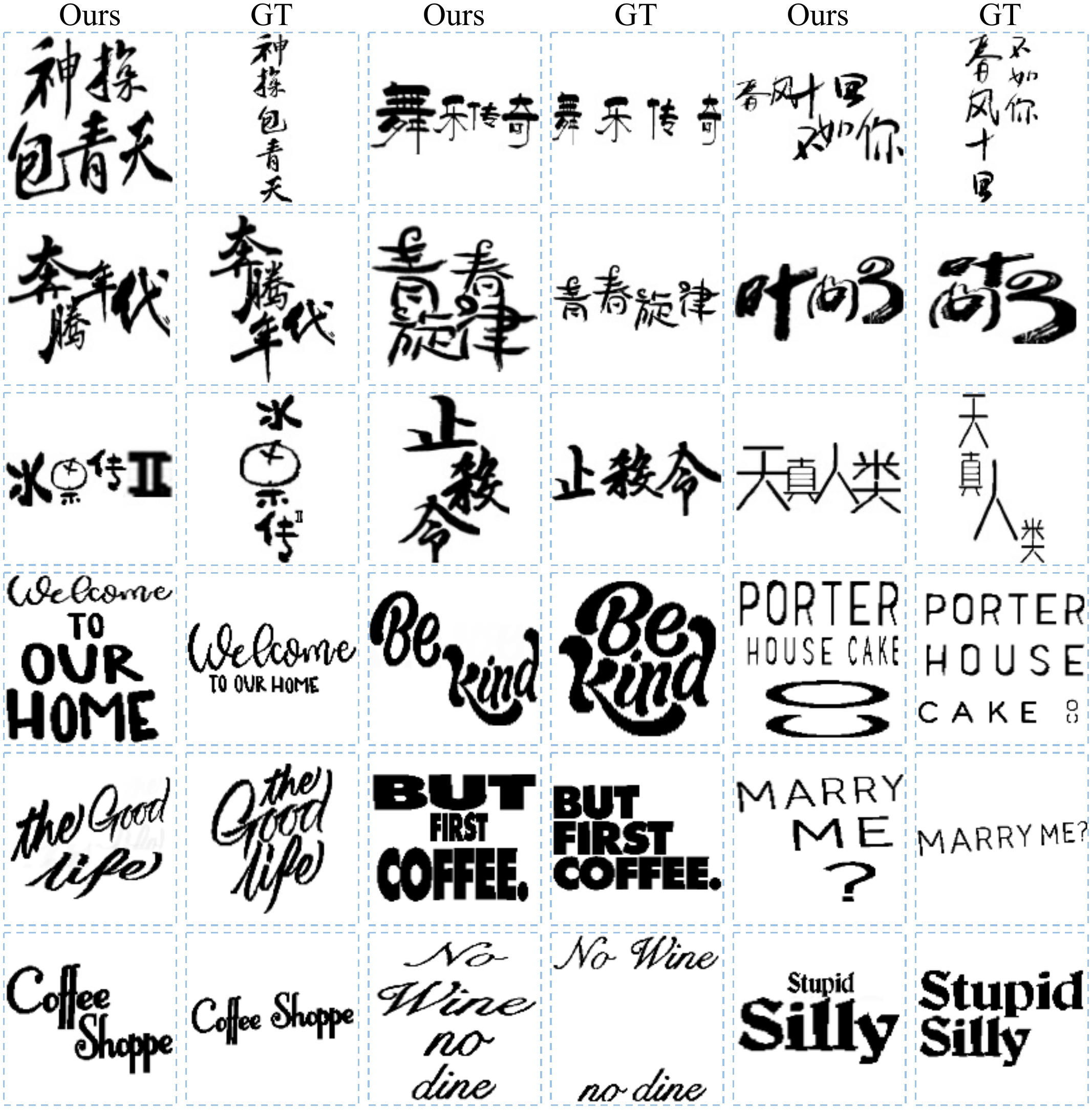}
  \vspace{-0.6cm}
  \caption{Our synthesized layouts and ground-truth (human-designed) layouts. ``GT" denotes ``ground truth".}
  \label{fig:SynExample}
\end{figure}

In Figure~\ref{fig:SynExample}, we demonstrate some text logos organized by our synthesized layouts and human-designed layouts.
The cases in the first three rows and the last three row are from TextLogo3K and TextSeg datasets, respectively.
According to the semantics of text, our model can generate line separation to split the tokens, such as the examples: and \begin{CJK}{UTF8}{gbsn}``神探/包青天"\end{CJK} (1st row, 1st case) and \begin{CJK}{UTF8}{gbsn}``春风十里/不如你"\end{CJK} (1st row, 3rd case).
Our model can also take the glyph shapes into consideration and generate compact layouts, such as the examples: \begin{CJK}{UTF8}{gbsn}``奔腾年代"\end{CJK} (2nd row, 1st case) and \begin{CJK}{UTF8}{gbsn}``天真人类"\end{CJK} (3rd row, 3rd case).
Our synthesis results for the English dataset are also visually pleasing, presenting layouts such as multiple new-lines, eye-catching scaling and translation.

\subsection{Random Sampling from Latent Space}
\begin{figure}[t!]
  \centering
  \includegraphics[width=\columnwidth]{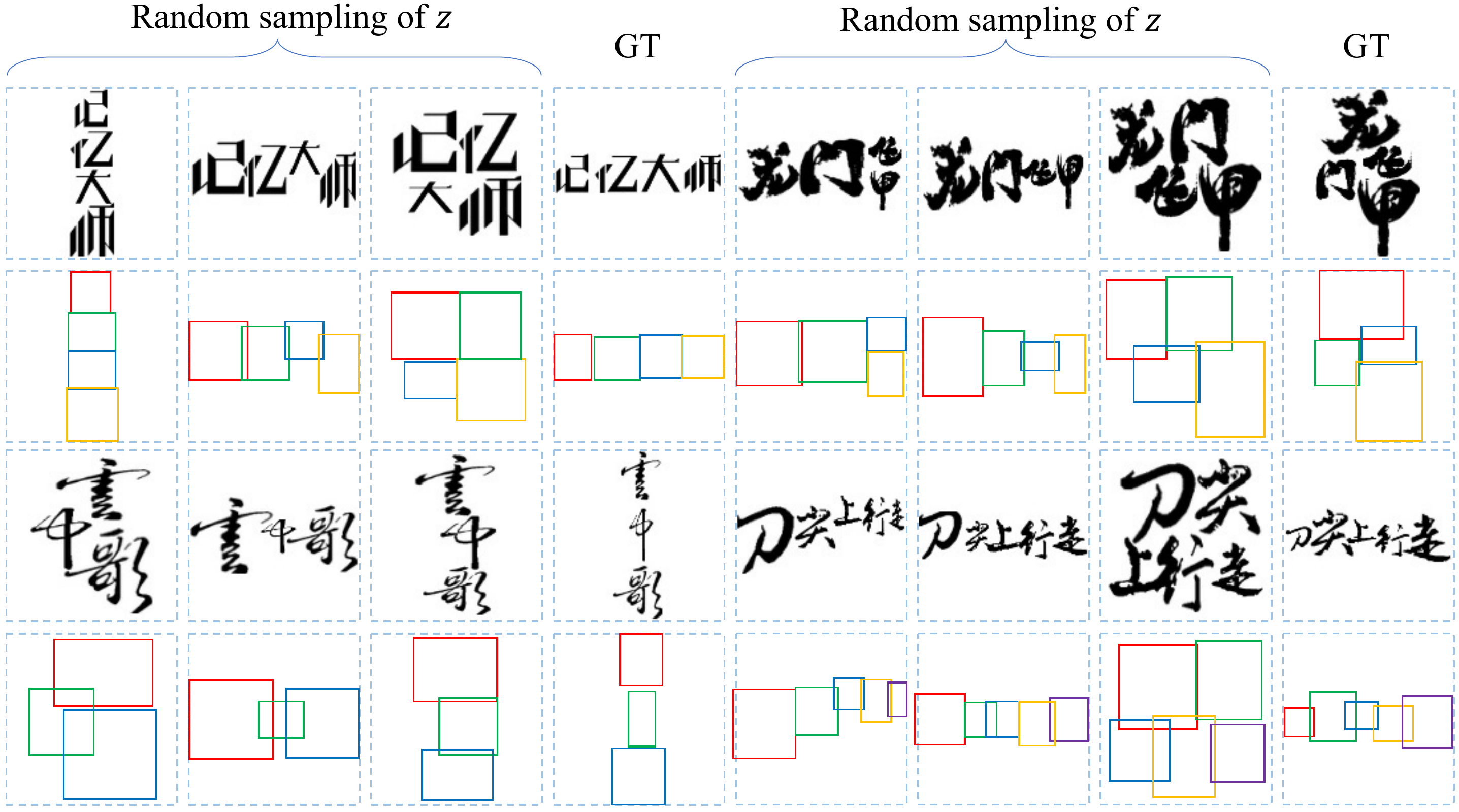}
  \vspace{-0.6cm}
  \caption{Synthesizing diverse layout styles by randomly sampling the latent noise $z$. Under each logo, the corresponding layout is shown by the bounding boxes of glyphs. A color sequence (from red to purple) is used to mark the order of different glyphs.}
  \label{fig:RandomNoise}
\end{figure}

As shown in Figure~\ref{fig:RandomNoise}, our model is capable of generating various styles given a set of inputs by randomly sampling $z$ for the Standard Normal Distribution.
The diverse styles of layouts include vertical texts, new lines, serpentine trajectories, etc.
The glyph shape information is carefully considered as we can see there is almost no overlap between different glyphs, although many of them have bounding boxes intersected.
The size of each character in a token is relatively consistent, such as \begin{CJK}{UTF8}{gbsn}``记忆"\end{CJK} (1st row, 1st case) and \begin{CJK}{UTF8}{gbsn}``龙门"\end{CJK} (1st row, 2nd case).
Featured with the diversity, the proposed model can provide designers with multiple choices of layouts and the designers can thus select their favourite one from them.
\begin{table}
	\centering
	\caption{The quantitative results of different methods and ablations of our method. ``Full'' denotes our full model. ``w/o Text'' and ``w/o Img'' denote that there are no text and no glyph image as the encoding input, respectively. $D_{s}$ and $D_{i}$ denote the Sequence Discriminator and the Image Discriminator, respectively. ``LoNet'' and ``LoGAN'' denote LayoutNet and LayoutGAN, respectively.}
	\vspace{-0.2cm}
	\begin{tabular}{lcccccc}
		\toprule
		 Ablation     & FID$\downarrow$     & IS $\uparrow$ &   Method     & FID$\downarrow$     & IS $\uparrow$ \\
		\midrule
         w/o Text  & 22.9  &  2.46 &  LoNet\cite{zheng2019content} & 35.6  & 2.13 \\
         w/o Img   & 29.0  &  2.39 &  LoGAN\cite{li2019layoutgan}   & 39.7  & 2.18  \\
         w/o $D_{s}$   & 55.6  & 1.98 & Rule (a) & 44.4  & 2.34  \\
         w/o $D_{i}$ & 22.5  & 2.42 & Rule (b) & 49.2  & 2.19  \\
         Full & \textbf{19.7} & \textbf{2.66} &  Rule (c) & 37.4  & 2.57   \\
		\bottomrule
	\end{tabular}
    \label{tab:ModelComparison}
\end{table}

\subsection{Ablation Studies}
\begin{figure}[t!]
  \centering
  \includegraphics[width=\columnwidth]{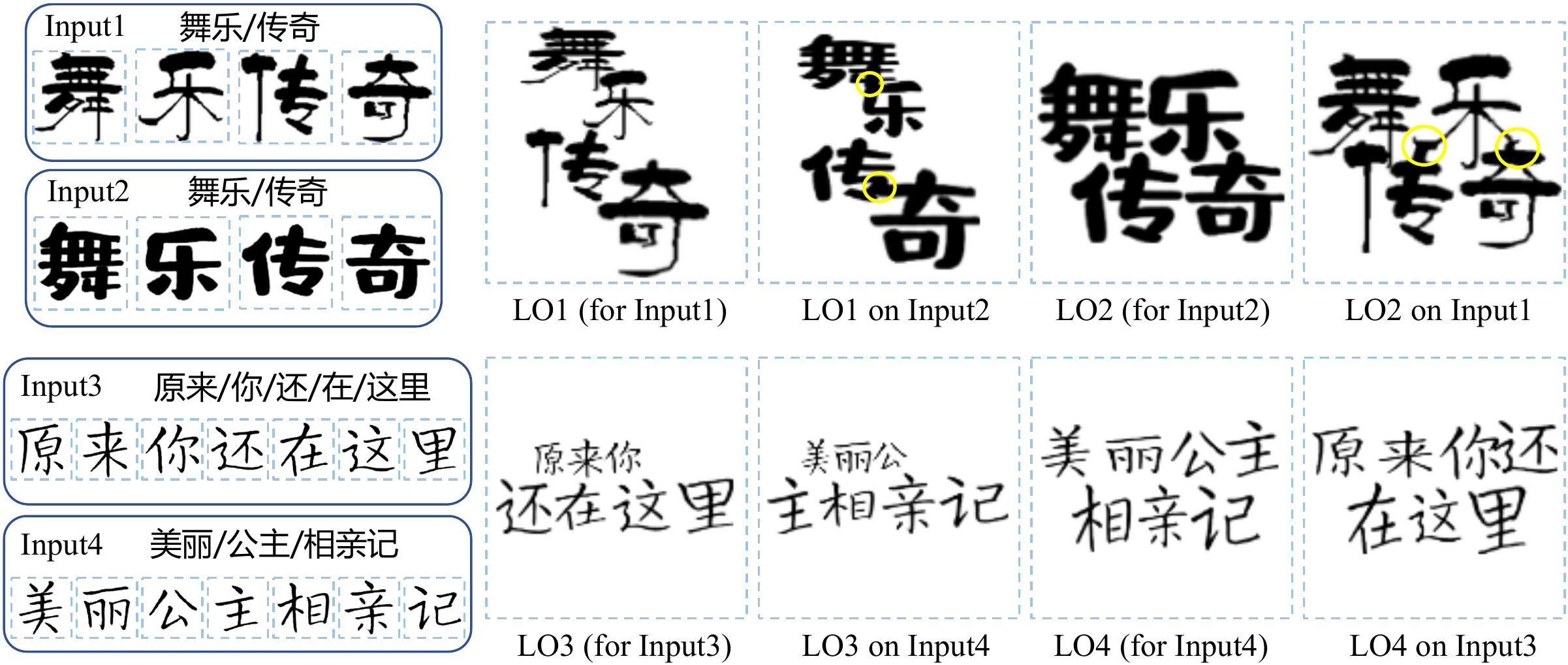}
  \vspace{-0.6cm}
  \caption{Testing our model's awareness of both visual and linguistic information. The yellow circles highlight the overlap between different glyphs. ``LO'' denotes layout. `/' is the symbol for spliting tokens.}
  \label{fig:InputTest}
\end{figure}

\begin{figure}[t!]
  \centering
  \includegraphics[width=\columnwidth]{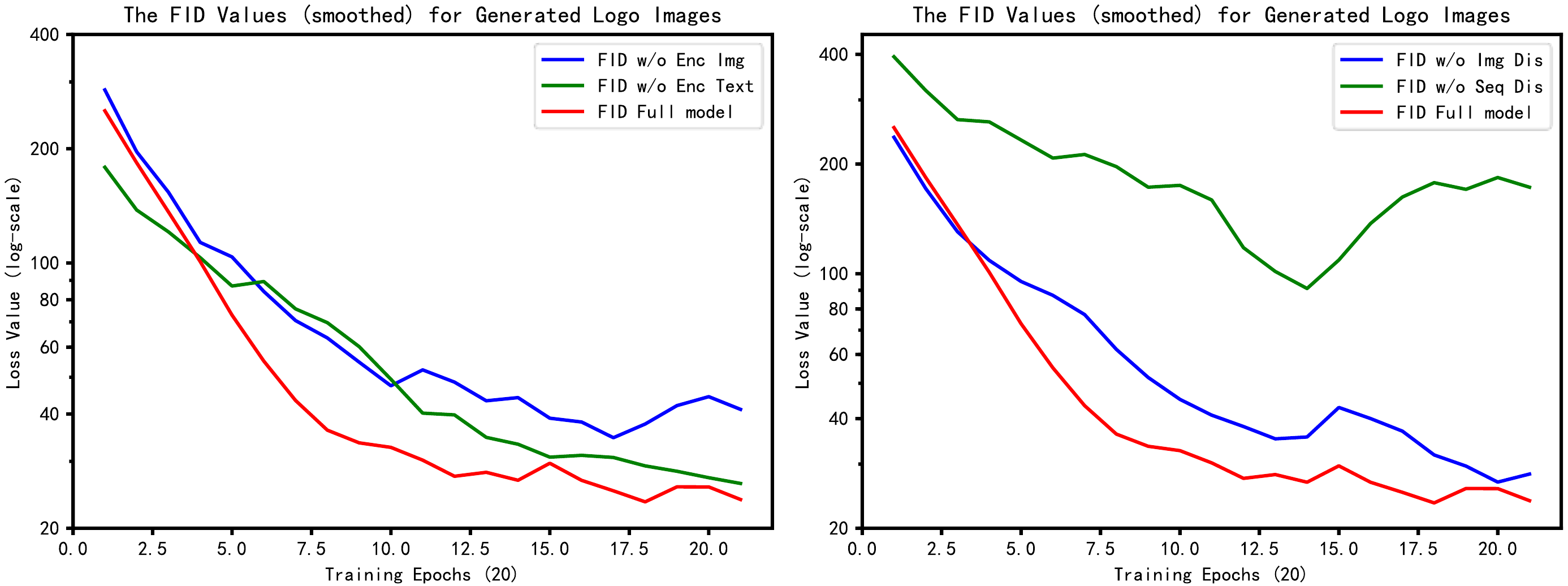}
  \vspace{-0.6cm}
  \caption{How FID values vary with training epochs under different ablations of our model. ``Enc Text" and ``Enc Img"  denote encoding the input text and images, respectively. ``Seq Dis" and ``Img Dis" denote the Sequence Discriminator and Image Discriminator, respectively.}
  \label{fig:AblationStudyQuan}
\end{figure}

\noindent\textbf{Effect of encoding both texts and images}.
As shown in Figure~\ref{fig:AblationStudyQuan}, the red curves are significantly lower than the blue and green curves, which indicates that better results can be achieved by (1) exploiting both visual and linguistic information of input elements (2) co-supervision from the Sequence Discriminator and Image Discriminator.
We find that the synthesized layouts usually do not conform with the text semantics if we do not encode the text linguistic information as inputs, e.g., starting a new line that breaks a token (provided in the supp. file).
On the other hand, if no glyph image is fed into the model, it can only infer the layouts by the linguistic information, the layouts might not be appropriate for each kind of glyph shape.
The FID and IS results with and without texts and images as input are reported in Table~\ref{tab:ModelComparison}, which also verify this conclusion.
To demonstrate our model's awareness of the visual and linguistic information, we send (1) the same text rendered in different font styles and (2) different texts in the same font style, respectively, into our model. Results are shown in Figure~\ref{fig:InputTest}, validating that our synthesized layouts are customized to the input texts and glyph images.
\par
\begin{figure}[t!]
  \centering
  \includegraphics[width=\columnwidth]{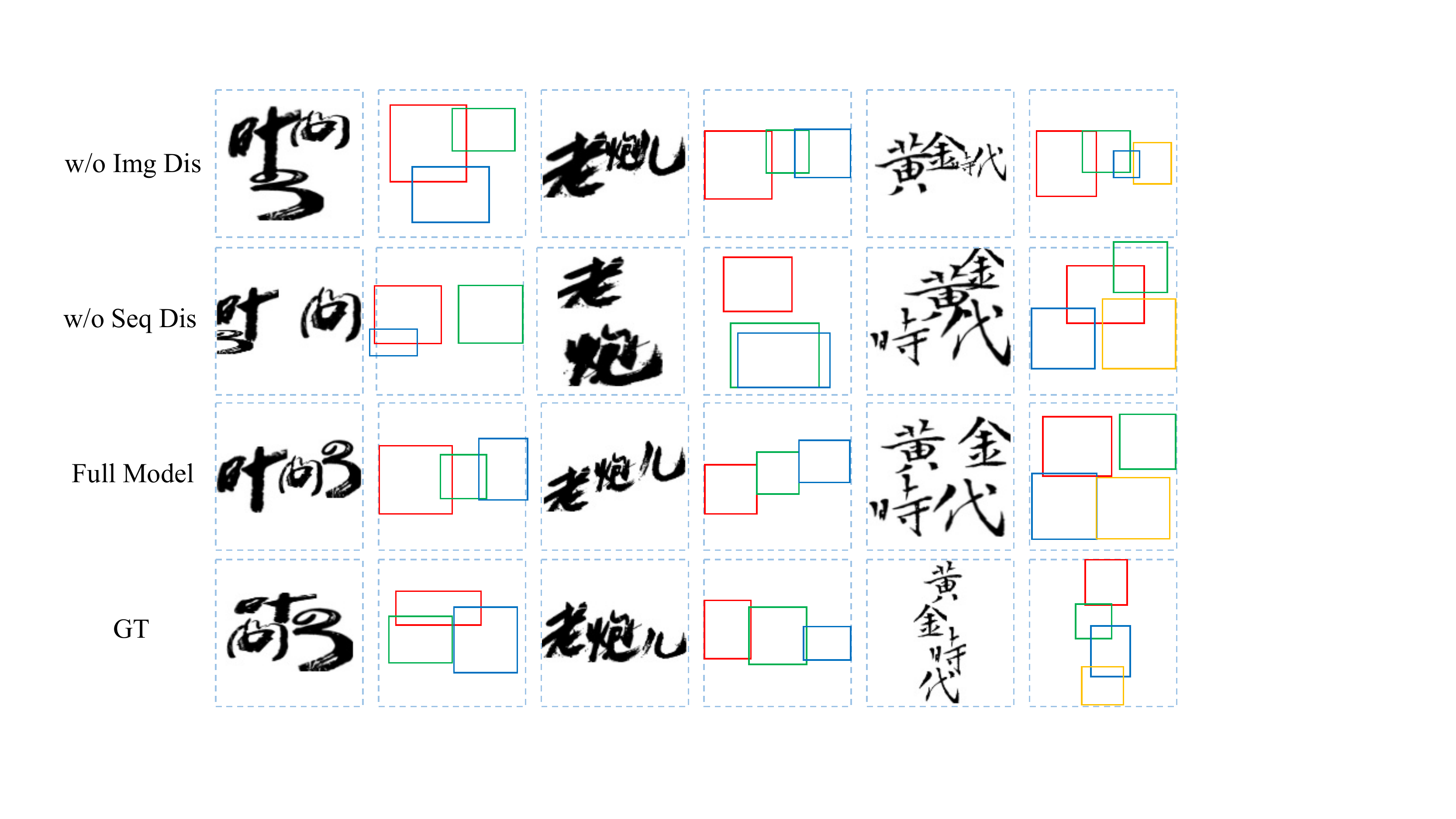}
  \vspace{-0.6cm}
  \caption{Qualitative results of the ablation studies. ``Img Dis" and ``Seq Dis" denote the Image and our Sequence Discriminator, respectively. The layout of each logo image is demonstrated by the bounding boxes next to it.}
  \label{fig:AblationStudyQual}
\end{figure}

\noindent\textbf{Effect of dual discriminators}.
As shown in Figure~\ref{fig:AblationStudyQual}, without the Sequence Discriminator, the module tends to generate abnormal sequences that do not conform to human reading habit.
For example, the `3' in \begin{CJK}{UTF8}{gbsn} in ``叶问3"\end{CJK} moves to the bottom-left of \begin{CJK}{UTF8}{gbsn}``叶"\end{CJK}, but the Image Discriminator cannot recognize it as a fake logo image.
Without the Image Discriminator, the details (such as spacing and stroke collision) are not as good as the results of the full model.
For example, the glyphs \begin{CJK}{UTF8}{gbsn} in ``老炮儿"\end{CJK} and  \begin{CJK}{UTF8}{gbsn} ``黄金时代"\end{CJK} are too close to each other.
Quantitative experimental results are shown in Table~\ref{tab:ModelComparison}, where removing the Sequence Discriminator (w/o Seq Dis) results in the worst performance.
\par

\noindent\textbf{Effect of the overlap loss}. We compute the values of $\mathcal{L}_{ol}$, FID, and IS on the testing dataset without and with optimizing $\mathcal{L}_{ol}$ when training the model. The results are shown in Table~\ref{tab:EffectofOL}, indicating that optimizing $\mathcal{L}_{ol}$  can significantly reduce the overlap and steadily improve FID and IS.
\begin{table}
	\centering
	\caption{The effect of the overlap loss. ``opt.'' denotes optimizing.  $\lambda \mathcal{L}_{ol}$ is evaluated on the testing dataset. }
	\vspace{-0.2cm}
	\begin{tabular}{lcccc}
		\toprule
		 Method  &  $\lambda \mathcal{L}_{ol}$ $\downarrow$   & FID$\downarrow$     & IS $\uparrow$  \\
		\midrule
         w/o opt. $\mathcal{L}_{ol}$  & 0.135  &  21.6 & 2.51  \\
         w/ opt. $\mathcal{L}_{ol}$   & 0.031  & 19.7 & 2.66 \\
		\bottomrule
	\end{tabular}
    \label{tab:EffectofOL}
\end{table}

\subsection{Comparison Against Other Methods}
\begin{figure}[t!]
  \centering
  \includegraphics[width=\columnwidth]{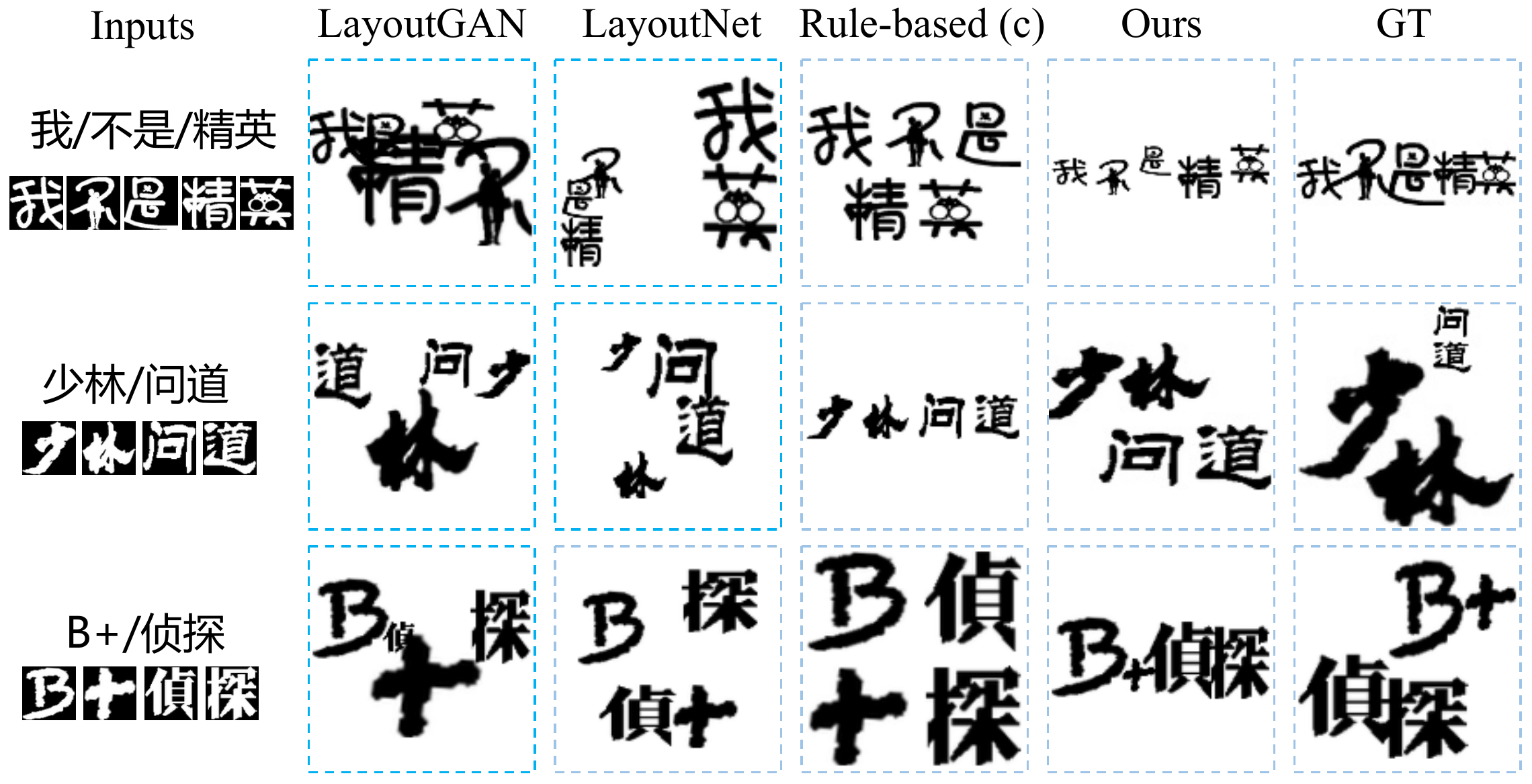}
  \vspace{-0.6cm}
  \caption{Qualitative comparison among different methods.}
  \label{fig:ComparisonQual}
\end{figure}

We compare our method with LayoutGAN~\cite{li2019layoutgan}, LayoutNet~\cite{zheng2019content}, and rule-based methods.
To make LayoutGAN and LayoutNet applicable in our task, the glyphs are simplified as rectangles and each character category is assigned with a corresponding color.
There are three rule-based methods tested in our experiments: (a) arranging all glyphs in a horizontal line; (b) arranging all glyphs in a horizontal or a vertical line (50\% probability); (c) arranging glyphs of one or more tokens (separated by Jieba\footnote{https://pypi.org/project/jieba/}) in a horizontal/vertical line, and randomly changing the spacing of different lines/glyphs.
From Table~\ref{tab:ModelComparison}, we can see that our method significantly outperforms others.
One major problem of LayoutGAN and LayoutNet is that they cannot generate layouts with correct reading orders, which is not fully reflected in Table~\ref{tab:ModelComparison} but vividly reflected in Figure~\ref{fig:ComparisonQual}.
This is because the image discriminator they both used can only capture the information such as characters' spacing but not their placing trajectories.
They also generate examples with more overlaps than ours, because they do not consider the detailed shapes of input elements.
The results also verify that the task of generating aesthetic text logos cannot be properly handled by rule-based methods, whose synthesized layouts are too regular and boring.

\subsection{User Study}
We conduct a Turing test where the participants (27 professional designers and 52 non-designers) are asked to select one from each pair of logos with the same text but different layouts (our synthesized and human-designed). 
For each logo pair, they are asked to choose (1) the logo image which they think is generated by machines (AI) (2) the logo image which they think has better quality. 
There are 20 examples with different layouts and font styles in this test. After finishing the Turing test, all participants are asked to give ratings (1 to 5, worst to best) for our synthesized layouts in terms of quality. 
The results are shown in Table~\ref{tab:UserStudy}, 
where the average accuracy of picking out our synthesized results is 54.6\% and the probability of our results being considered better is 40.8\%, which indicates that the quality of our synthesized layouts is comparable to human-designed ones.
\begin{table}
	\centering
	\caption{Results of the user study conducted to evaluate our synthesized layouts. The ratings of ``Quality'' are from 1 to 5.}
	\vspace{-0.2cm}
	\begin{tabular}{lccc}
		\toprule
		User Group & Accuracy$\downarrow$ & Prefer.$\uparrow$ & Quality$\uparrow$  \\
		\midrule
		Designers & 58.1\%   &  33.0\%   &  3.0  \\
		Non-designers & 52.7\% &  44.9\%  & 3.8    \\
		Average     &  54.6\%  & 40.8\% & 3.5  \\ 

		\bottomrule
	\end{tabular}
    \label{tab:UserStudy}
\end{table}

\subsection{Text Logo Synthesis System}
\begin{figure}[t!]
  \centering
  \includegraphics[width=\columnwidth]{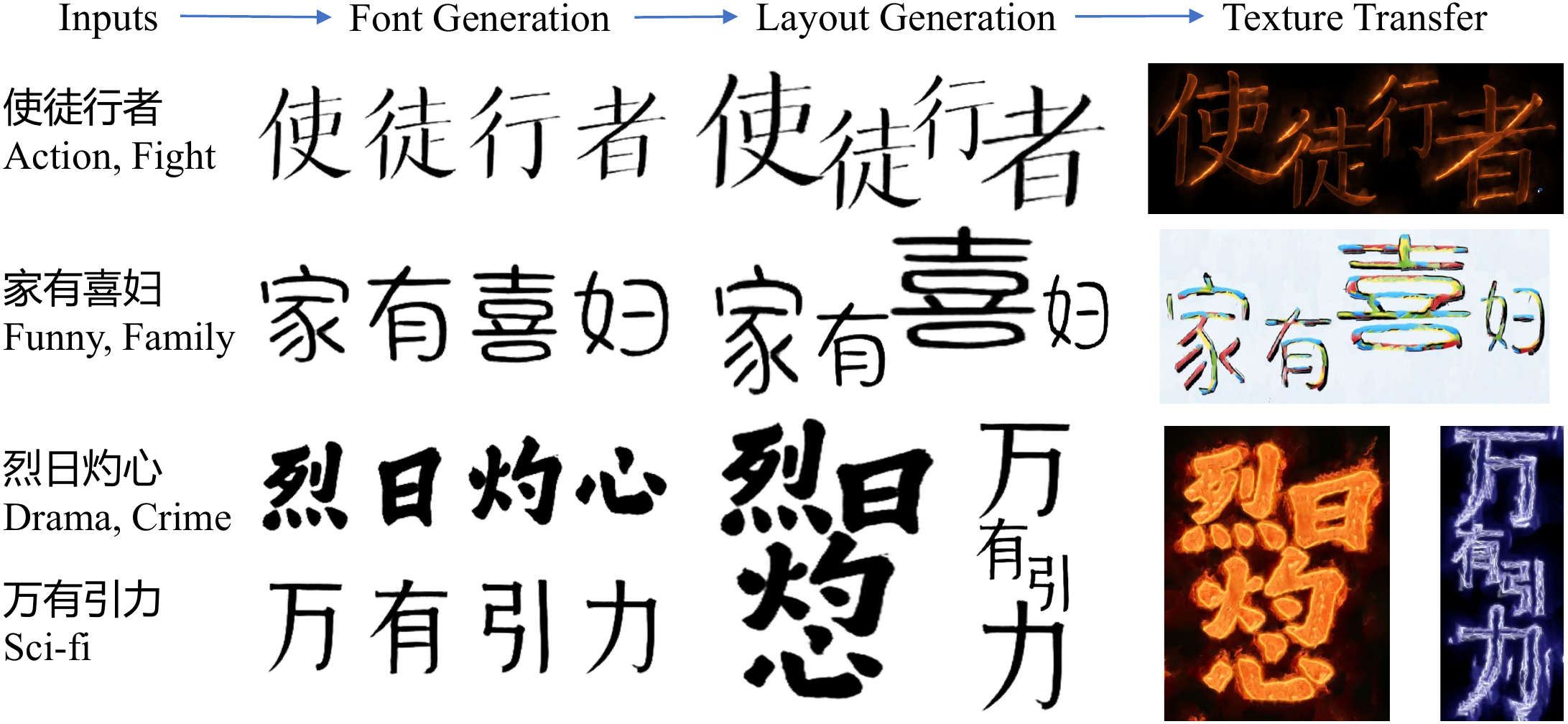}
  \vspace{-0.6cm}
  \caption{Synthesizing text logos by combing our layout generation method with font generation and texture transfer approaches.}
  \label{fig:SystemDemo}
\end{figure}

Inspired by font generation and texture transfer models (\cite{zhang2018separating,gao2019artistic,wang2020attribute2font}, and \cite{yang2017awesome,men2018common,men2019dyntypo}), we build a text logo synthesis system which takes a desired text and its topic as input and synthesizes the text logo image automatically.
Specifically, we first synthesize the font according to the user-input topics, such as lovely, funny, technical, etc.
Implementation details about the font synthesis network are described in the supplementary file.
Then, we send the synthesized glyph images into our layout generation network to synthesize the text logo image.
Finally, we perform the texture transfer method proposed by~\cite{men2019dyntypo} to obtain the final text logo with texture effects. Several examples are shown in Figure~\ref{fig:SystemDemo}, demonstrating the effectiveness of our system for synthesizing aesthetic text logos.

\subsection{Limitations}
\begin{figure}[t!]
  \centering
  \includegraphics[width=\columnwidth]{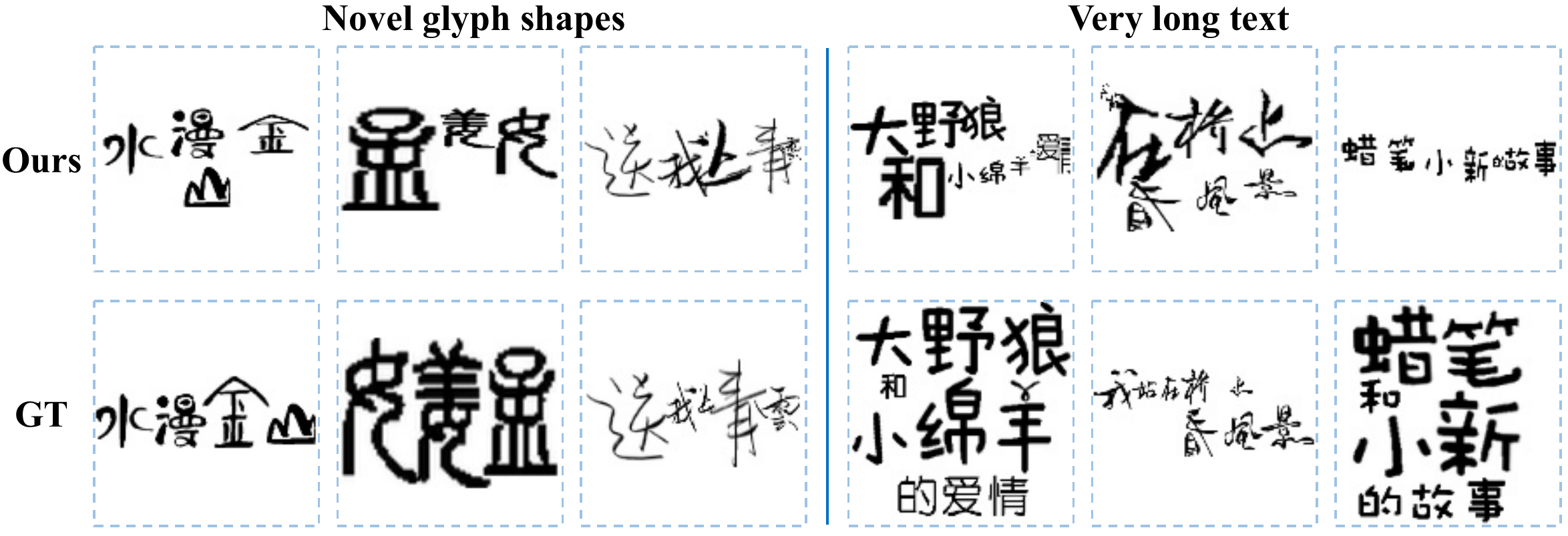}
  \vspace{-0.6cm}
  \caption{Some failures cases of our model.}
  \label{fig:Limitations}
\end{figure}

Some failure cases of our method are shown in Figure~\ref{fig:Limitations}.
The proposed layout generation network is not robust enough to glyphs with novel shapes, which could affect our model's understanding of the text semantics since the image features are concatenated with the character embeddings.
Besides, currently our model is still unable to satisfactorily handle texts with too many characters/words ($N > 8 $), failing to arrange appropriate positions for some glyphs.

\section{Conclusion}
In this paper, we proposed a layout generation network for aesthetic text logo synthesis.
Both the linguistic and visual information of input texts and glyph images were taken into account for layout prediction. Moreover, a novel dual-discriminator module was designed to capture both the placing trajectories and detailed shapes of characters for synthesizing high-quality layouts.
We built a large-scale dataset, TextLogo3K, on which extensive experiments (both quantitative and qualitative) were conducted to verify the effectiveness of our method. 
Finally, a prototype system was developed to automatically synthesize aesthetic text logos by combing our layout generation network with existing font generation and texture transfer models.

\section*{Acknowledgements}
This work was supported by Beijing Nova Program of Science and Technology (Grant No.: Z191100001119077), Project 2020BD020 supported by PKU-Baidu Fund, Center For Chinese Font Design and Research, Key Laboratory of Science, Technology and Standard in Press Industry, State Key Laboratory of Media Convergence Production Technology and Systems.

{\small
\bibliographystyle{ieee_fullname}
\bibliography{egbib}
}

\end{document}